\documentclass[letterpaper, 10 pt, conference]{ieeeconf}  % Comment this line out if you need a4paper

\IEEEoverridecommandlockouts                              % This command is only needed if 
\usepackage{xcolor}
\usepackage{hyperref}
\usepackage{graphicx}

\usepackage[english]{babel}
\usepackage[T1]{fontenc}

\usepackage[nolist]{acronym}
\usepackage[cmex10]{amsmath}
\usepackage[font={small}]{caption}
\usepackage{subcaption}
\usepackage[export]{adjustbox}
\usepackage{amssymb}
\usepackage{bm}
\usepackage{booktabs}
\usepackage{booktabs}
\usepackage{balance}
\usepackage{cite}
\usepackage{color}
\usepackage{dsfont}
\usepackage{esvect}
\usepackage{float}
\usepackage{graphicx}
\usepackage{hyperref}
\usepackage{import}
\usepackage{mathtools}
\usepackage{multirow}
\usepackage{multicol}
\usepackage{multirow}
\usepackage{flushend}
\usepackage{outline}
\usepackage{paralist}
\usepackage{siunitx}
\usepackage{stmaryrd}
\usepackage{systeme}
\usepackage{soul}
\usepackage{url}
\usepackage{tikz}
\usetikzlibrary{tikzmark}
\usepackage[normalem]{ulem}

\newcommand{\vxcmd}{\textbf{v}_x^{\footnotesize{\textrm{cmd}}}}
\newcommand{\vycmd}{\textbf{v}_y^{\footnotesize{\textrm{cmd}}}}

\newcommand{\projectwebsite}{\url{https://gmargo11.github.io/dribblebot}}

\newcommand{\simvideo}{Video S1}

\title{\LARGE \bf
DribbleBot: Dynamic Legged Manipulation in the Wild
}

\author{Yandong Ji$^*$, Gabriel B. Margolis$^*$, and Pulkit Agrawal% <-this % stops a space
\thanks{$^*$ indicates equal contribution.}% <-this % stops a space
\thanks{All authors are with the Improbable AI Lab,
        Massachusetts Institute of Technology, USA. Correspondence to:
        {\tt\footnotesize \{yandong, gmargo\}@mit.edu}}%
}

\begin{document}

\let\oldtwocolumn\twocolumn
\renewcommand\twocolumn[1][]{%
    \oldtwocolumn[{#1}{
    \begin{flushleft}
           \centering
    \includegraphics[clip,trim=0cm 0cm 0cm 0cm,width=0.95\textwidth]{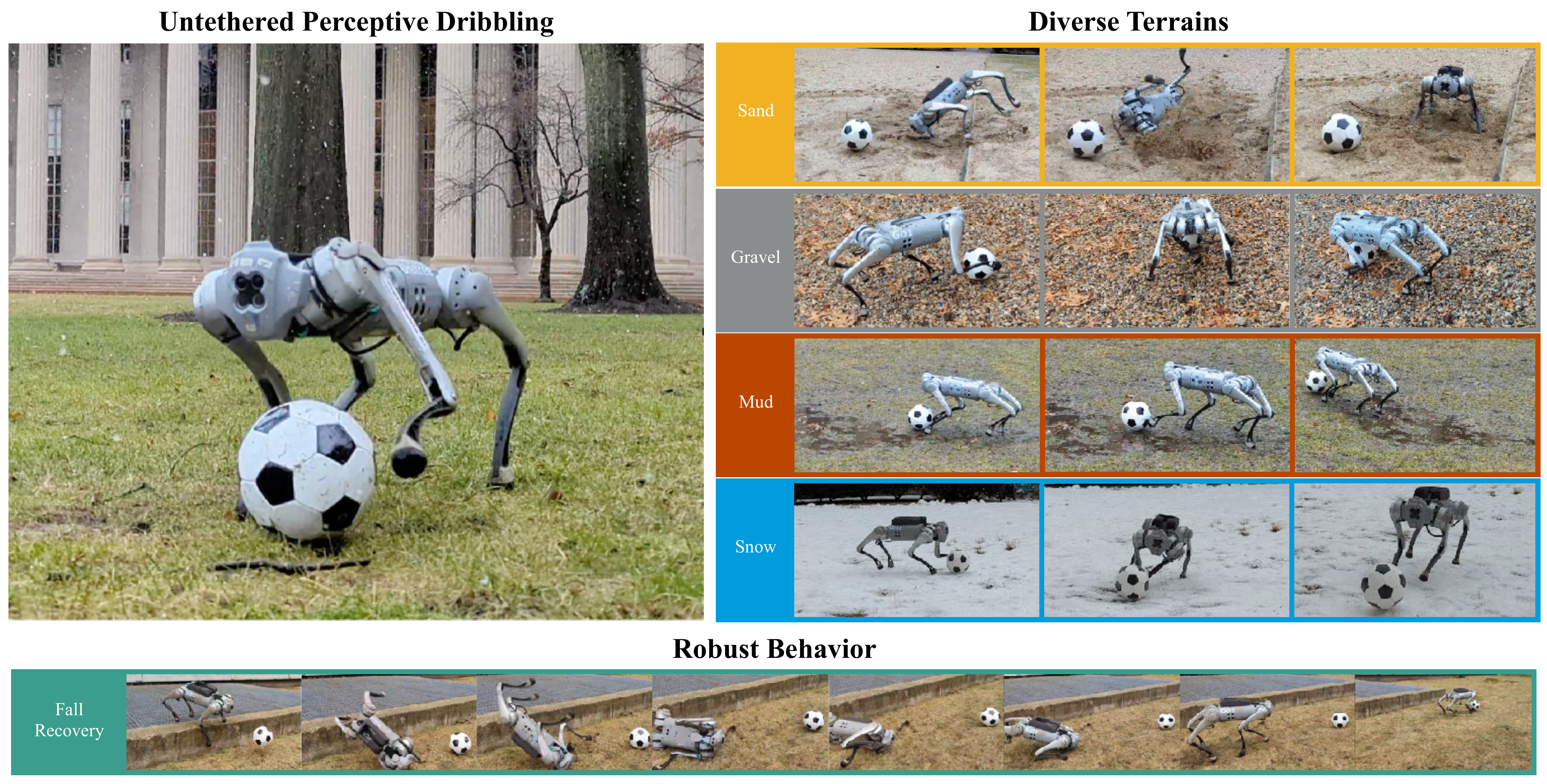}
    \captionof{figure}{
    \textbf{\textit{In-the-wild} dribbling} on diverse natural terrains including sand, gravel, mud, and snow using onboard sensing and computing. Our system trained using reinforcement learning successfully adapts to varying ball dynamics on different terrains, and can get up and recover the ball after falling down from an extreme perturbation.
    }\label{fig:setup}
    \end{flushleft}
    }]
}

\maketitle
\thispagestyle{empty}
\pagestyle{empty}

%%%%%%%%%%%%%%%%%%%%%%%%%%%%%%%%%%%%%%%%%%%%%%%%%%%%%%%%%%%%%%%%%%%%%%%%%%%%%%%%
\begin{abstract}

DribbleBot (\underline{D}exte\underline{r}ous \underline{B}all Manipulation with a \underline{Le}gged Ro\underline{bot}) is a legged robotic system that can dribble a soccer ball under the same real-world conditions as humans (i.e., \textit{in-the-wild}). We adopt the paradigm of training policies in simulation using reinforcement learning and transferring them into the real world. We overcome critical challenges of accounting for variable ball motion dynamics on different terrains and perceiving the ball using body-mounted cameras under the constraints of onboard computing. Our results provide evidence that current quadruped platforms are well-suited for studying dynamic whole-body control problems involving simultaneous locomotion and manipulation directly from sensory observations. Video and code are available at \projectwebsite.

\end{abstract}

% \newpage

%%%%%%%%%%%%%%%%%%%%%%%%%%%%%%%%%%%%%%%%%%%%%%%%%%%%%%%%%%%%%%%%%%%%%%%%%%%%%%%%

\section{INTRODUCTION}
Consider dynamic mobile manipulation, a family of compound tasks requiring tight integration of visual perception, dynamic locomotion, and object manipulation. In applications like package delivery or search-and-rescue, a robot is often required to move quickly while carrying an object. Instead of the object being held at a fixed position in the robot's reference frame, due to time constraints, it may be necessary to manipulate the object while the robot is moving. 
For instance, a recent demonstration showcased a humanoid Atlas robot picking up, running with, and throwing heavy objects in a construction site~\cite{deits2023picking}.
Past studies of dynamic mobile manipulation have been limited due to the requirement of expensive hardware and the complexity of simultaneously optimizing locomotion and manipulation objectives while estimating the state of the robot, object, and environment, which is even more challenging in dynamic scenarios. Consequently, prior work has primarily focused either on individual subproblems of locomotion and manipulation or on a less dynamic version of the combined problem. 
However, in the past few years, legged robot hardware equipped with visual and proprioceptive sensors has become more accessible. Reinforcement learning has emerged as a powerful paradigm for addressing complex contact-rich problems such as robust legged locomotion across challenging terrains like stairs, hiking trails, sand, and mud~\cite{tan2018sim,hwangbo2019learning,lee2020learning, kumar2021rapid,siekmann2021blind,miki2022learning,margolis2022rapid,margolis2022walktheseways,agarwal2022legged,choi2023learning} and dynamic object manipulation using dexterous hands~\cite{andrychowicz2020learning,akkaya2019solving, chen2021system,chen2022visual}. This work investigates whether we can extend the same approach to dynamic mobile manipulation tasks. 

As a case study, we consider the task of dribbling a soccer ball in the wild, which requires task-oriented coordination of all the robot's joints (\textit{whole-body control} \cite{khatib2004whole}) 
%\pulkit{citation needed}
to kick and pursue the ball while also maintaining balance. A human athlete 
operates from onboard perception and can dynamically control a ball on a wide variety of natural terrains, including grass, mud, snow, and pavement. 
In contrast, previous studies on legged robot soccer 
operated under one or more following restrictions: (i) use of flat and smooth playing surface that does not require dealing with terrain variation \cite{veloso1998playing, stone2007intelligent, friedmann2008versatile, leottau2015study, bohez2022imitate, ji2022soccer, huang2022creating}; (ii) use of multiple static external cameras in indoor settings which bypasses the need to deal with noisy onboard visual perception and lighting variation~\cite{bohez2022imitate, ji2022soccer, huang2022creating}; (iii) not performing locomotion and ball-kicking simultaneously, but rather pushing the ball with the body~\cite{veloso1998playing, stone2007intelligent, friedmann2008versatile} or executing static dribbling~\cite{leottau2015study, bohez2022imitate, ji2022soccer} where the ball comes to rest before each kick.
We present a method that overcomes these limitations. 

Bringing soccer dribbling from the laboratory into the wild is not merely a matter of synthesizing techniques from locomotion and manipulation but presents several unique challenges. One is adapting to the ball-terrain dynamics, which varies independently from the robot-terrain dynamics due to the light weight of the ball and the different nature of rolling contact compared to the contact of the feet against the ground. 
On pavement, the ball may roll away faster than the robot can run; on grass, the ball will slow down quickly and requires more frequent and stronger kicks. We overcome this challenge by augmenting the simulator with a custom ball drag model. 
Another consideration is the limited precision and range of onboard cameras. When a small robot is dribbling close to its body, it is hard to localize the ball using a conventional body-mounted camera due to its narrow field of view. Instead, we use observations from multiple wide-angle fisheye cameras, for which we develop a separate ball detection module to facilitate sim-to-real transfer. 
Finally, the robot can experience locomotion failure on challenging terrains that are outside the training distribution of the dribbling controller, such as a steep curb.  
To address these scenarios, we integrate a recovery policy trained for harsher conditions that enables the robot to stand up autonomously after a fall. We find this controller can help to regain control of the ball and continue to dribble.

The resulting system, DribbleBot (\underline{D}exte\underline{r}ous \underline{B}all Manipulation with a \underline{Le}gged Ro\underline{bot}), demonstrates dynamic real-world dribbling maneuvers across a variety of terrains. By providing evidence that existing hardware and sensors are capable of successful behavior, we hope to motivate more work in both robot soccer and, more generally, on the problem of dynamic mobile manipulation. 

\section{MATERIALS}
\textbf{Hardware}: We use the Unitree Go1 robot~\cite{unitree2022website} and a size $3$ soccer ball for all experiments. This small robot quadruped stands \SI{40}{\centi\meter} tall. We use two onboard $210^{\circ}$ field-of-view fisheye cameras to capture images, one facing forward and one facing downward. All computation is performed on two onboard NVIDIA Jetson Xavier NX units. Due to the computation, communication bandwidth, and electrical power limitations of the robot, we process full-resolution images locally on each Jetson unit and send only the ball location estimates to the computer running the policy.

\textbf{Simulator}: We simulate the Unitree Go1 robot in Isaac Gym~\cite{makoviychuk2021isaac} using the manufacturer-provided URDF model. Simulation and training run on a single NVIDIA RTX 3090.

\textbf{Object Detection Module}: We use YOLOv7~\cite{redmon2016you, wang2022yolov7} model weightspretrained on the COCO dataset~\cite{lin2014microsoft} and finetune it to perform ball detection as described in section \ref{sec:yolo}. We accelerate object detection
using TensorRT allowing processing of images with resolution $400 \times 480$ pixels comfortably at \SI{30}{\hertz}.

\section{METHOD}

\begin{figure}[t!]
    \centering
    \includegraphics[width=\linewidth]{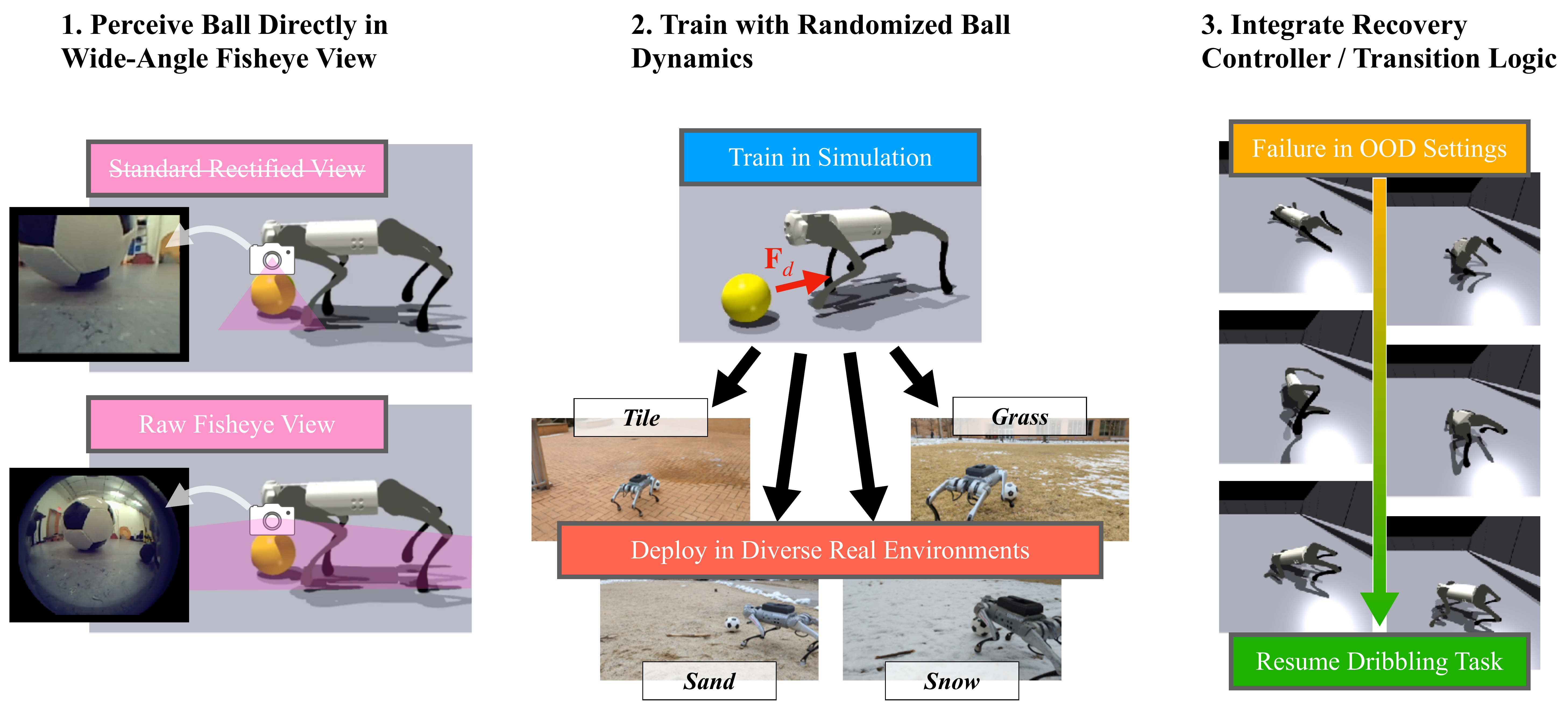}
    \caption{\textbf{Measures added for sim-to-real transfer in DribbleBot.}}
    \label{fig:sim2real_measures}
\end{figure}

\textbf{Overview}: We train control policy $\mathbf{a}_t = \pi_d(\mathbf{o}_t, \mathbf{c}_t)$ in simulation and transfer it to the real world. The observation $\mathbf{o}_t$ consists of the proprioceptive data and the ball position $\mathbf{b}_t$. The ball velocity command $\mathbf{c}_t$ is provided as input (e.g., by a human user during deployment). We choose not to train our policy end-to-end on camera images because of the numerous challenges, including slow simulation speed, poor sample efficiency in training from high-dimensional visual observations, and the sim-to-real gap in image observations. Instead, we train the policy using ball position that is easily available in simulation. For real-world deployment, a separately trained object detection model ($\mathbf{Y}$) predicts the ball position from images $(\mathbf{o}^v_t)$ captured by on-board cameras: $\hat{\mathbf{b}}_t = \mathbf{Y}(\mathbf{o}^v_t)$. The policy $\pi_d$ outputs actions $\mathbf{a}_t$, which are the joint position targets of twelve motors (three motors per leg) at \SI{50}{\hertz}. $\pi_d$ is trained using reinforcement learning algorithm, Proximal Policy Optimization (PPO)~\cite{schulman2017proximal}. 

\subsection{Training the Dribbling Policy}
\subsubsection{Environment Design} We train the robot in simulation to dribble the ball on flat ground with physical parameters randomly varied as detailed in Section \ref{sec:noise_model}. At the start of every episode, the robot's yaw orientation is randomly initialized, and its initial leg positions are randomized around a nominal pose. The soccer ball is initialized at a random position within \SI{2}{\meter} of the robot. The target ball velocity is also uniformly randomized. These considerations ensure that the robot learns omnidirectional locomotion and dribbling. The episode length is \SI{40}{\second}, and the control timestep is \SI{50}{Hz}.

\subsubsection{Control Interface for Dribbling in the Wild}
\label{sec:control-interface}
Successful dribbling involves adjusting the leg swings to apply targeted forces while the robot moves, balances itself, and orients its position relative to a moving ball. Previous works in sim-to-real legged locomotion commonly use body velocity command or position~\cite{rudin2022advanced} as the control interface, with a few exceptions that allow the user to tweak gait~\cite{siekmann2021simtoreal, margolis2022walktheseways} or foot placement~\cite{duan2022sim} parameters.  The dribbling skill is not easily expressed just with gait or body-level control commands, and issuing such commands would be unintuitive for a human user deploying the robot. Instead, we directly command the ball's linear velocity in the 2D ground plane expressed in the global reference frame. Because the ball has full rotational symmetry and the robot's orientation can vary rapidly during a kicking maneuver, the local reference frames of both the robot and the ball are constantly changing and less useful for a human operator. We choose to command the robot in the global frame as it does not change with the robot's motion making it much easier for the human to control the robot. The local body frame of the robot at the first time step serves as the global frame of reference. 

\begin{figure}[t!]
    \vspace{0.2cm}
    \centering
    \includegraphics[clip,trim=0cm 0cm 0cm 0cm,width=0.95\linewidth]{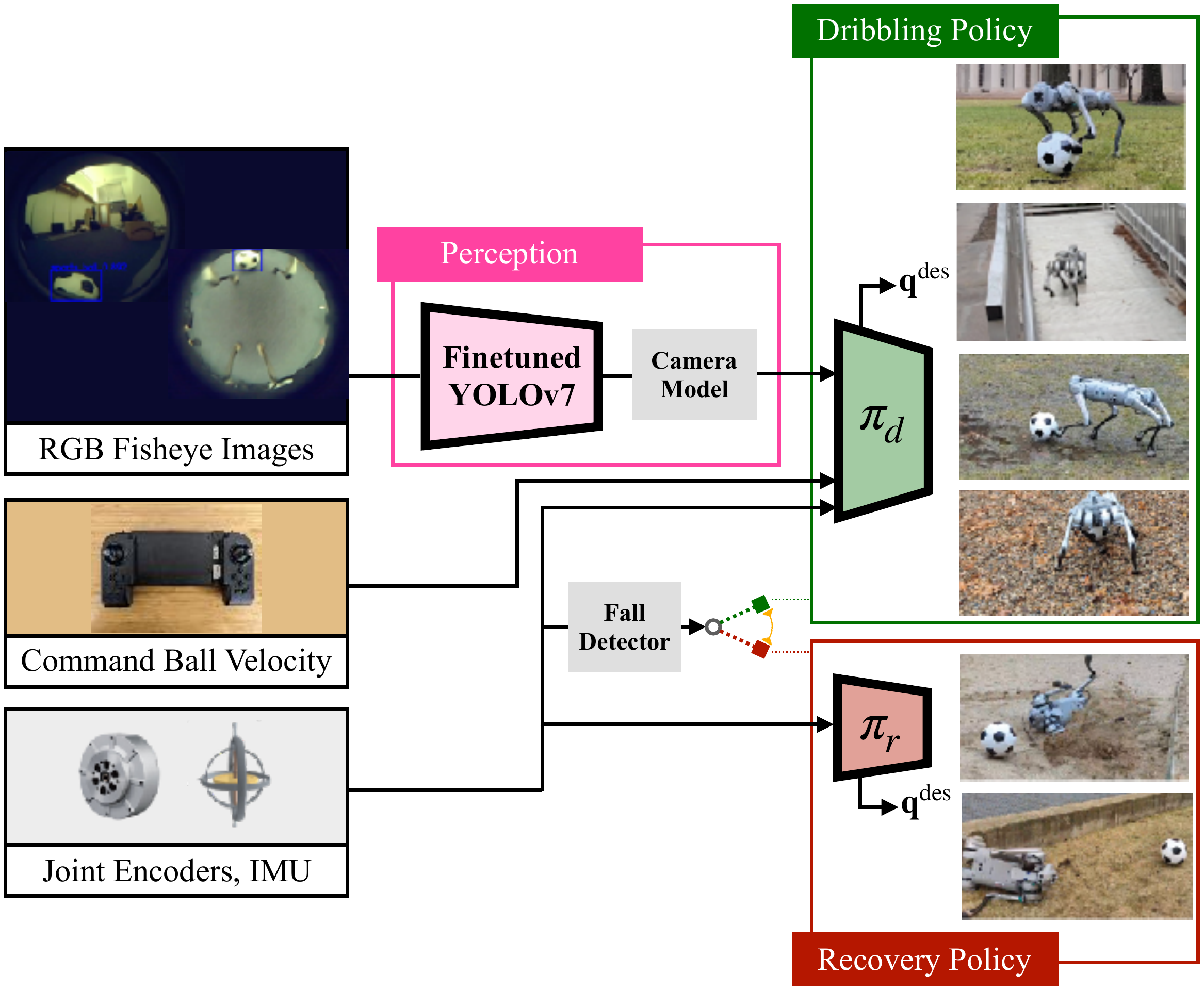}
    \caption{
   \textbf{ System architecture for DribbleBot.} $\pi_d$ and $\pi_r$ are multilayer perceptrons trained using reinforcement learning in simulation. YOLOv7 is an object detection network~\cite{wang2022yolov7} that we fine-tune on images from our domain using supervised learning.}\label{fig:architecture}
\end{figure}

\subsubsection{Observation and Action Space} 
\label{sec:observation-action-space}
The input to the policy $\pi_d$ is $\mathbf{o}_t$ consisting of the $15$-step history of command $\mathbf{c}_t$, ball position $\mathbf{b}_t$, joint positions and velocities $\mathbf{q}_t, \dot{\mathbf{q}}_t$, gravity unit vector in the body frame $\mathbf{g}_t$, global body yaw $\pmb{\psi}_t$, and timing reference variables $\pmb{\theta}^\mathrm{cmd}_t$ as defined in in~\cite{margolis2022walktheseways}.  The commands $\mathbf{c}_t$ consist of the target ball velocities ${\vxcmd, \vycmd}$ in the global frame (Sec.~\ref{sec:control-interface}). For the robot to process commands in the global frame, it must know its orientation in the global frame for which we provide the global body yaw $\pmb{\psi}_t$ obtained from the IMU as input to the policy. 
The action space $\mathbf{a}_t$ is the target position of the twelve joints maintained by a high-frequency PD controller with $k_p=20.0$, $k_d=0.5$~\cite{margolis2022walktheseways}.

\subsubsection{Reward Model}
\label{sec:reward-model}
Table \ref{tab:reward_table} (Appendix) provides the task reward terms used for ball dribbling. Task rewards include a reward for tracking the commanded ball velocity in the global reference frame and a reward shaping term incentivizing the robot to be close to the ball. Since the camera has a limited observation range, to promote ball visibility in the camera, the robot is also rewarded for facing the ball.
A set of gait reward terms~\cite{siekmann2021simtoreal, margolis2022walktheseways} encourage the robot to adopt a consistent ground contact schedule and generate well-formed gaits.  
Additional standard safety reward terms~\cite{rudin2021learning, margolis2022rapid, ji2022concurrent, margolis2022walktheseways} are included to penalize dangerous commands and facilitate the sim-to-real transfer. 

\subsubsection{Policy Architecture and Optimization}
We use Proximal Policy Optimization (PPO)~\cite{schulman2017proximal} to train our soccer dribbling policy represented as a neural network with three fully-connected hidden layers of sizes $[512, 256, 128]$.
The policy converges after $7$ billion timesteps, or about $48$ hours of training. For visualization and debugging, it is ideal to have access to state variables: the robot's body velocity, ball velocity, and ball-terrain drag force coefficient. 
A regression model represented as a two-layer neural network with $[256,128]$ units is trained in simulation to predict these parameters~\cite{ji2022concurrent} from $\mathbf{o}_t$ (Sec.~\ref{sec:observation-action-space}). Additionally, the input to the policy network is augmented with these predicted parameters. While prior works found this technique 
useful for sim-to-real transfer of running policies~\cite{ji2022concurrent}, we did not evaluate its effect on sim-to-real performance in the dribbling task. 

\subsection{Measures to Mitigate the Sim-to-Real Gap}
\subsubsection{Visual Perception in Fisheye Images} 
\label{sec:yolo}
DribbleBot needs to localize a size-$3$ soccer ball (diameter \SI{18}{\centi\meter}) using a body-mounted camera when the ball is as close as \SI{10}{\centi\meter} to the robot. This precludes the use of standard depth cameras like RealSense due to their narrow field of view (${105^\circ}$). Such cameras can only see a small part of the ball when it is close -- not enough to localize it (Figure \ref{fig:sim2real_measures}). 
Instead, our robot is equipped with ultrawide forward-facing and downward-facing fisheye cameras, each with a field of view $210^{\circ}$.  
Rather than simulating these images during policy learning, which would be slow and engender a large sim-to-real gap, we construct a separate ball-detection module based on YOLOv7~\cite{wang2022yolov7}, an object detection model pre-trained on a large image dataset~\cite{lin2014microsoft}). 
With such a wide view, rectification results in substantial warping of the spherical soccer ball, so the off-the-shelf performance of models trained on datasets of narrow-view rectified images taken from the internet is not as good. 
However, we find that good detection performance can be easily recovered in the entire field of view by fine-tuning YOLOv7 on 254 hand-labeled fisheye images of soccer balls from our robot's camera, including images with the ball at the edge. We apply standard data augmentation to the images, including horizontal flip, rotations up to \SI{15}{\degree}, and blur. 

Object detection outputs a bounding box, but our control policy takes as input the ball position. We obtain ball position from monocular RGB images through an approximate application of the equidistant fisheye lens model. 
Given the ball pixel coordinates, we first compute the angle $\theta$ from the camera principal axis to the ball center using the equidistant model $r=f\theta$ where $r$ is the distance from the image center in pixels and $f$ is the camera focal length. 
Ignoring warping effects, we calculate the size of the ball in the image in pixels $\Delta r$ from its bounding box, 
and knowing the physical ball radius $R$, we again apply the 
equidistant fisheye model 
to compute the approximate distance between the ball and the camera $z$: $z = f\frac{R}{\Delta r}$. Knowing the distance from the camera to the ball and the direction from the camera to the ball determines the ball's position in the camera frame. We transform the observed ball position into the body frame using the known camera extrinsics. 
Finally, if the ball is detected in both front and bottom cameras, a single position estimate is obtained by selecting the detection with a higher confidence value.

\subsubsection{Vision Noise Model}
While in simulation, accurate ball position estimates are available, in the real world, the ball position estimates are noisy. To ensure that the policy is robust to visual perception noise, we add noise sampled from a uniform distribution to the ball position during training. Further, to emulate large changes in the ball position that might happen due to a human kicking the ball or the ball going outside the robot's field of view, we also randomly teleport the ball in the ground plane. 
Finally, because the data rate of the camera is limited, we simulate camera communication delay. Noise model details are provided in Section \ref{sec:noise_model} (Appendix).

\subsubsection{Robot System Identification} 
To mitigate the sim-to-real gap in robot dynamics, we employ two standard and effective system identification measures: (a) Train an actuator network on real-world torque data to account for the non-ideal motor dynamics~\cite{lee2019robust, margolis2022walktheseways,chen2022visual}; and (b) Identify and model the lag between the time the observation is measured and the time the action is applied~\cite{xie2021dynamics, margolis2022walktheseways,chen2022visual}.

\subsubsection{Ball-Terrain Interaction Model}
The variable drag force resulting from rolling surface contact between a soccer ball and different terrains is not modeled by the rigid-body simulator we used.
Human soccer players can quickly adapt to variations in ball dynamics due to physical factors like air pressure and size as well as external perturbations to the ball due to uneven terrain or opposing players. We embed robustness in DribbleBot by implementing a custom domain randomization scheme. 
Our domain randomization includes a ball drag model that applies drag force proportional to the square of the velocity: $F_D = C_D v^2$, following the standard equation for aerodynamic drag. Different values of $C_D$ serve to emulate terrains with various resistance forces, such as a field with tall grass (high $C_D$) or pavement (low $C_D$). In addition, we randomize the ball mass and apply random changes in ball velocity during training. Ball velocity randomization simulates external perturbations to the ball, such as human intervention or contact with uneven terrain. The randomization range details are given in Table \ref{tab:domain_rand} (Appendix).

\subsubsection{Fall Recovery Controller} 
If the robot encounters a harsh perturbation, such as a shove or a steep curb, and experiences locomotion failure, this will also cause a loss of ball control. In this scenario, we would like the robot to get up from its fall and resume dribbling.
Similar to prior works~\cite{lee2019robust, smith2022legged}, we train a dedicated recovery policy that enables the robot to return to a standing position from diverse fall scenarios. The recovery policy is trained in harsher domains, including uneven terrain and randomized gravity force, scenarios that are too challenging for the dribbling controller. We first generate a set of 1000 initial fall configurations by randomly dropping the robot from different orientations, then train a policy with rewards for body orientation, base height, and action smoothness. The details of the reward function and training procedure for the recovery policy are provided in Table \ref{tab:reward_table} (Appendix).
To transition between dribbling and recovery policy, we define a finite state machine with transitions based on the body orientation. When the roll or pitch angle is larger than \SI{1.0}{rad}, indicating locomotion failure, the recovery policy executes a return to the standing pose.  When roll and pitch are smaller than \SI{0.5}{rad}, the dribbling policy is reactivated.

\section{RESULTS}

\subsection{Simulation Performance}

\begin{table}[t!]
\centering
\vspace{0.22cm}
\bgroup
\def\arraystretch{1.5}
{
\begin{tabular}{lrcclrc}
\cline{1-3}
\cline{5-7}
\multicolumn{3}{|c|}{Tile} & & \multicolumn{3}{|c|}{Grass}  \\
\cline{1-3}
\cline{5-7}
\texttt{Full} & $4$/$4$ & \multirow{3}{*}{\parbox{0.20\linewidth}{ \centering \vspace{0.1cm} \includegraphics[width=\linewidth,clip,trim={0cm 0 0cm 0}]{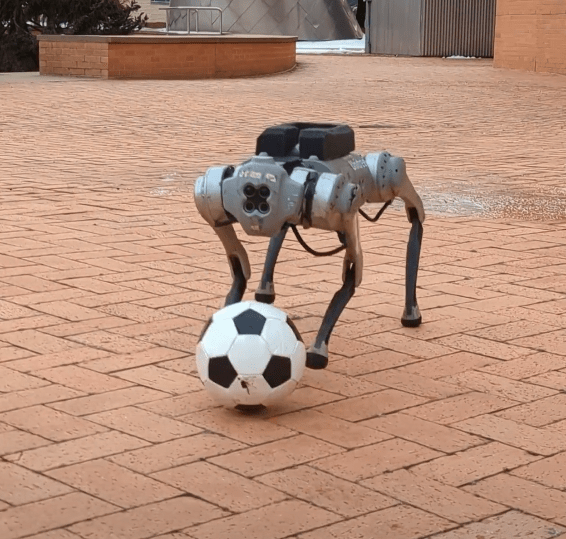}}} &  & \texttt{Full} & $4$/$4$ & \multirow{3}{*}{\parbox{0.20\linewidth}{ \centering \vspace{0.1cm} \includegraphics[width=\linewidth,clip,trim={0cm 0 0cm 0}]{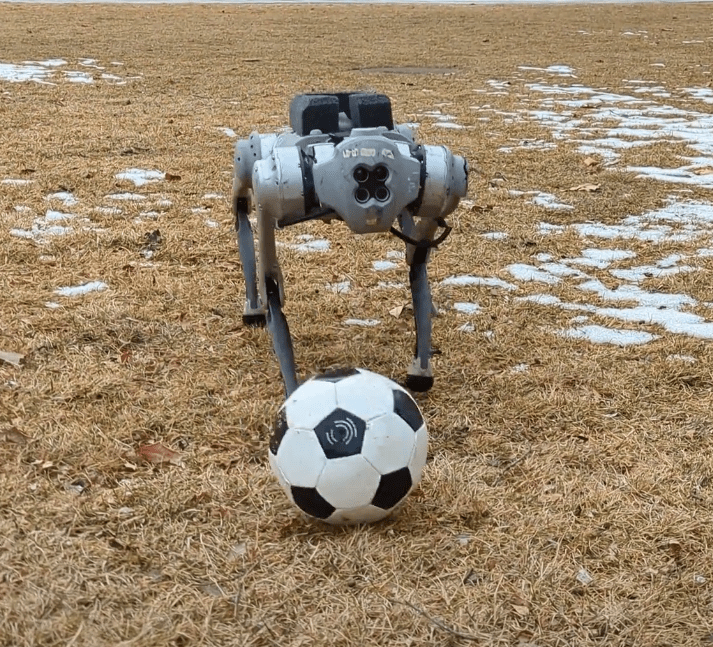}}} \\
\texttt{-R} & $4$/$4$ & & & \texttt{-R} & $4$/$4$ \\
\texttt{-Y} & $0$/$4$ & & & \texttt{-Y} & $0$/$4$  \\
\texttt{-D} & $4$/$4$ & & & \texttt{-D} & $4$/$4$  \\
 % & & \\
\cline{1-3}
\cline{5-7}
\multicolumn{3}{|c|}{Sand} & & \multicolumn{3}{|c|}{Snow}  \\
\cline{1-3}
\cline{5-7}
\texttt{Full} & $4$/$4$ & \multirow{3}{*}{\parbox{0.20\linewidth}{ \centering \vspace{0.1cm} \includegraphics[width=\linewidth,clip,trim={0cm 0 0cm 0}]{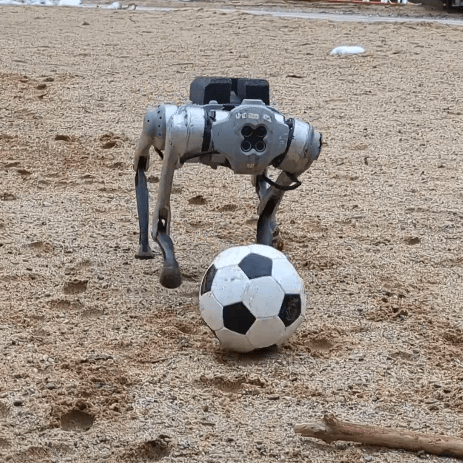}}} &  & \texttt{Full} & $3$/$4$ & \multirow{3}{*}{\parbox{0.20\linewidth}{ \centering \vspace{0.1cm} \includegraphics[width=\linewidth,clip,trim={0cm 0 0cm 0}]{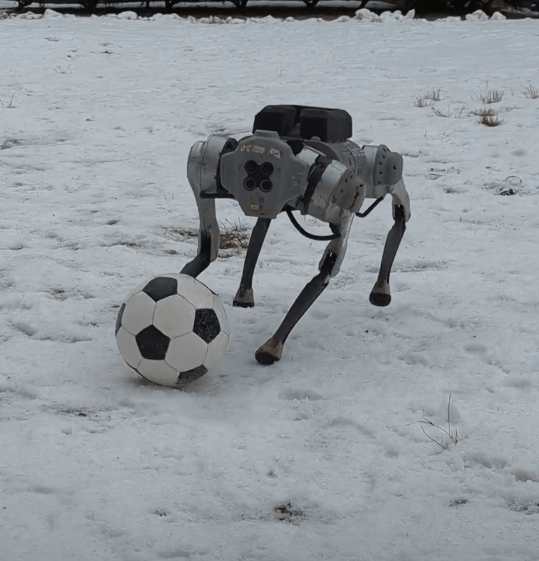}}} \\
\texttt{-R} & $4$/$4$ & & & \texttt{-R} & $3$/$4$ \\
\texttt{-Y} & $0$/$4$ & & & \texttt{-Y} & $0$/$4$  \\
\texttt{-D} & $2$/$4$ & & & \texttt{-D} & -  \\
\cline{1-3}
\cline{5-7}
\multicolumn{3}{|c|}{Curb Step-Down} & & \multicolumn{3}{|c|}{Ramp}  \\
\cline{1-3}
\cline{5-7}
\texttt{Full} & $2$/$4$ & \multirow{3}{*}{\parbox{0.20\linewidth}{ \centering \vspace{0.1cm} \includegraphics[width=\linewidth,clip,trim={4cm 0 3cm 0}]{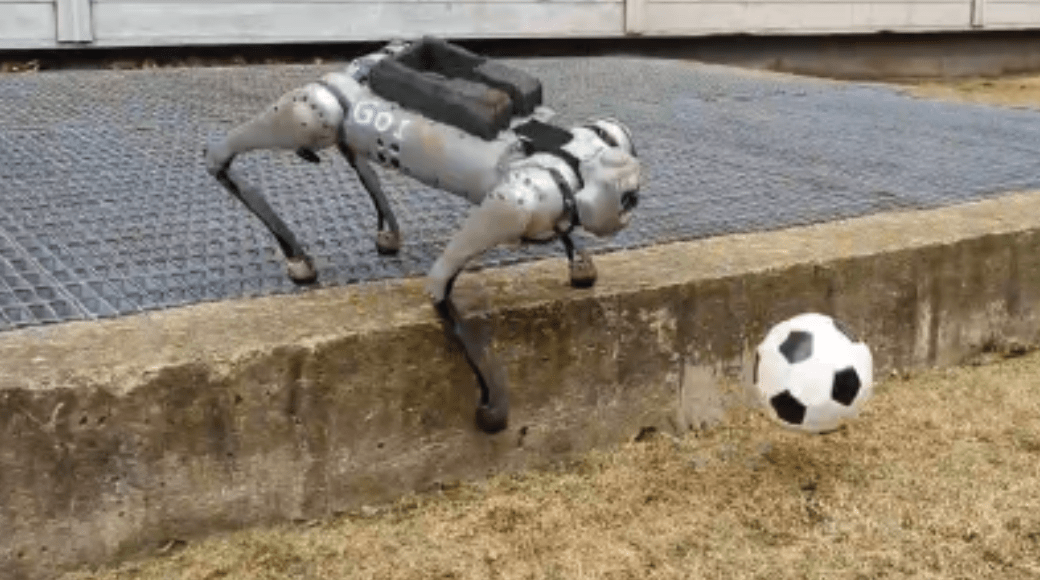}}} &  & \texttt{Full} & $0$/$4$ & \multirow{3}{*}{\parbox{0.20\linewidth}{ \centering \vspace{0.1cm} \includegraphics[width=\linewidth,clip,trim={2cm 0 5cm 0}]{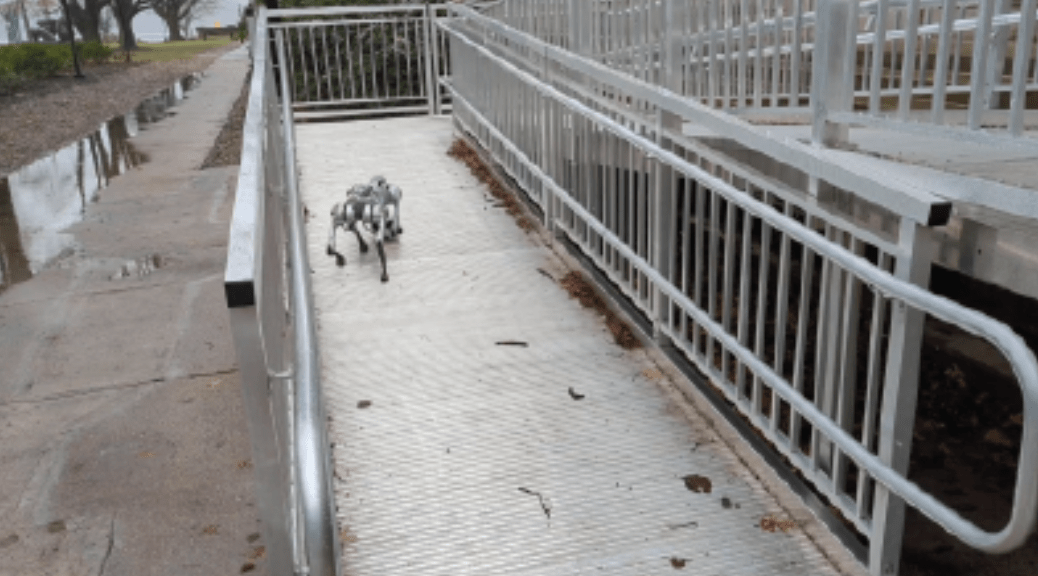}}} \\
\texttt{-R} & $1$/$4$ & & & \texttt{-R} & $0$/$4$ \\
\texttt{-Y} & - & & & \texttt{-Y} & -  \\
\texttt{-D} & - & & & \texttt{-D} & -  \\
\hline
\end{tabular}
}
\egroup
\caption{\textbf{Real-world dribbling performance evaluation.} The robot executes a fixed dribbling trajectory on each trial. We test each scenario with full system design (\texttt{Full}) and with ablations: No recovery controller (\texttt{-R}); No YOLO fine-tuning (\texttt{-Y}); No drag model during training (\texttt{-D})
}\label{tab:evaulation_result}
\end{table}

\newdimen\figrasterwd
\figrasterwd\textwidth

\begin{figure*}[t]
    \vspace{0.2cm}
    \centering
  \adjustbox{minipage=0.0em}{\subcaption{}\label{sfig:path_viz}}%
  \begin{subfigure}[t]{\dimexpr\linewidth}%-1.3em\relax}
  \centering
  \includegraphics[clip,trim=0cm 0cm 0cm 0cm,width=.8\linewidth,valign=t]{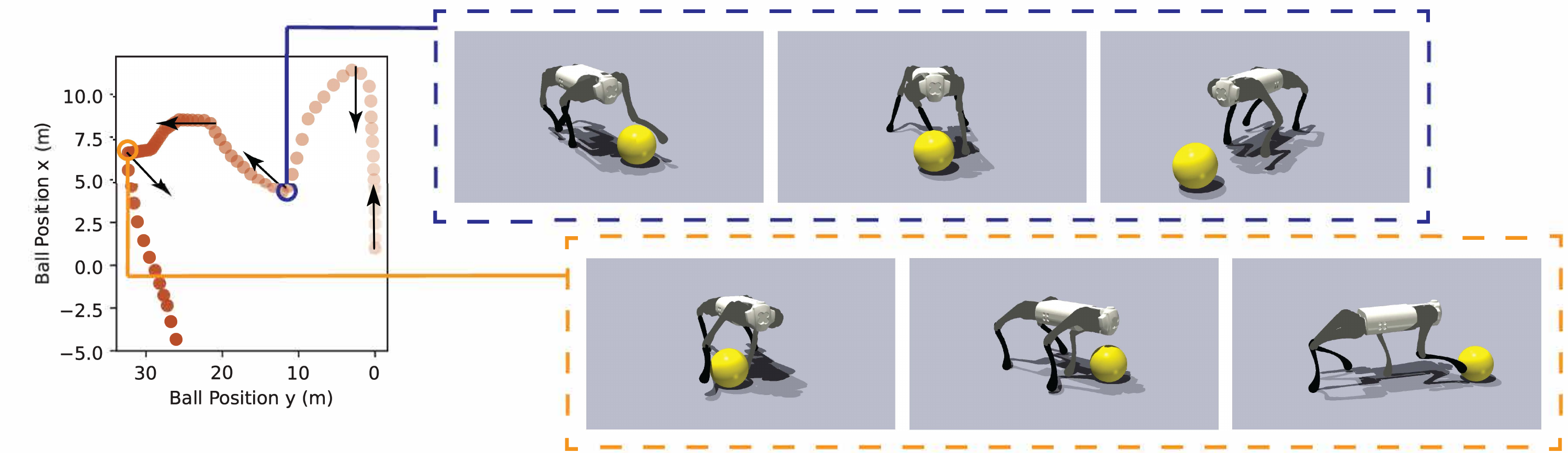}
  \end{subfigure}%
  \hfill
  \adjustbox{minipage=0.0em}{\subcaption{}\label{sfig:ang_vel}}%
  \begin{subfigure}[t]{\dimexpr\linewidth}%-1.3em\relax}
  \centering
  \includegraphics[clip,trim=0cm 0.1cm 0cm 0cm,width=.8\linewidth,valign=t]{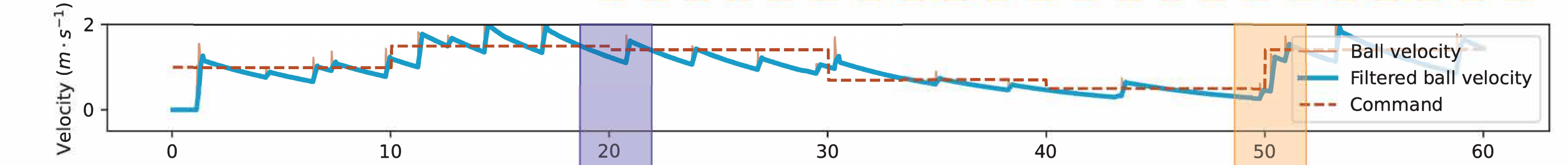}
  \end{subfigure}%
  \hfill
  \adjustbox{minipage=0.0em}{\subcaption{}\label{sfig:lin_vel}}%
  \begin{subfigure}[t]{\dimexpr\linewidth}%-1.3em\relax}
  \centering
  \includegraphics[clip,trim=0cm 0.1cm 0cm 0.1cm,width=.8\linewidth,valign=t]{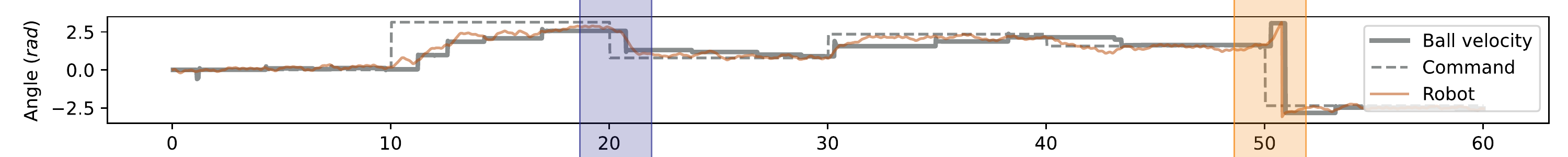}
  \end{subfigure}%
  \hspace{0.0cm}
  \adjustbox{minipage=0.0em}{\subcaption{}\label{sfig:joints}}%
  \begin{subfigure}[t]{\dimexpr\linewidth}%-1.3em\relax}
  \centering
  \includegraphics[clip,trim=0cm 0cm 0cm 0cm,width=.8\linewidth,valign=t]{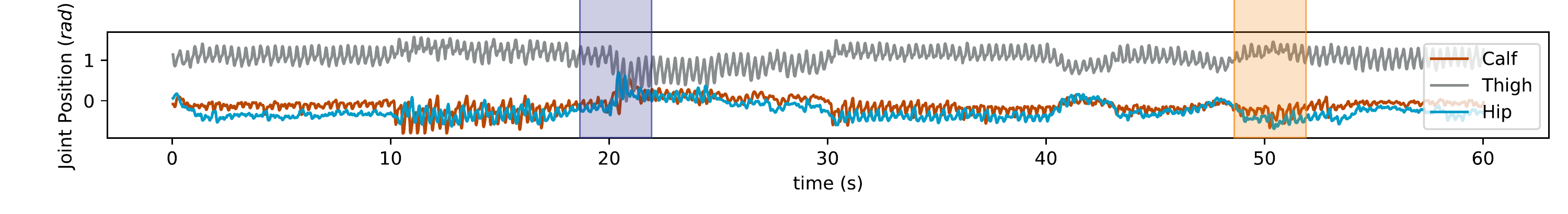}
  \end{subfigure}
    \caption{
   \textbf{ Simulated dribbling performance evaluation.} (a) Ball position in world frame and turning moment snapshots. The red points indicate the ball's position and darken as time elapses. The dark arrows represent the approximate direction of the commanded velocity. 
   {(b), (c)} Ball velocity tracking performance in polar coordinates. 
   Here, the robot is first commanded to dribble the ball forward at \SI{1}{m/s} and then to execute a sequence of turns at various speeds.
   (d) We illustrate the joint position of the front right leg, which is typically used for dribbling and for executing left turns.
   The blue highlight around \SI{20}{\second} corresponds to the left turn visualized in the first row of images above. The orange highlight around \SI{50}{\second} corresponds to the right turn visualized in the second row. 
    }
    \vspace{-0.2cm}
    \label{fig:drib_sim_perf}
\end{figure*}

\subsubsection{Dribbling Control} 
We first evaluate our dribbling policy in simulation under the same conditions experienced during training. If the robot is dribbling well, we can characterize its performance by how closely it tracks the ball velocity command, how fast it can dribble, and how sharply it can turn.
Figure \ref{fig:drib_sim_perf} visualizes a long \SI{60}{\second} run and shows the dribbling path, command tracking performance, and joint motion. The corresponding behavior is shown in \simvideo.

In the simulated experiment, the robot is able to track dribbling speeds up to \SI{1.5}{m/s} and the entire range of instantaneous changes in command direction up to $180^{\circ}$. Unlike in ordinary locomotion, the act of changing the ball velocity may extend across a long time horizon of several seconds, during which the robot establishes control of the ball and executes multiple kicks
(Figure \ref{sfig:ang_vel}, \ref{sfig:lin_vel}). 

We observe that dynamically dribbling a soccer ball in this manner requires task-oriented coordination of all the robot's joints (\textit{whole-body control}). 
Figure \ref{sfig:joints} shows that dribbling a soccer ball involves different participation of the right front leg during left and right turns. This leg is used to kick the ball during a left turn (blue highlight, Figure \ref{sfig:path_viz} upper images), and its motion is substantially changed during this maneuver. Later, when the robot makes a right turn (orange highlight, Figure \ref{sfig:path_viz} lower images), the left front leg is used to kick the ball, but the motion of the right leg also adjusts to stabilize the maneuver.

\subsection{Real-world Deployment}

\subsubsection{Qualitative Results on Diverse Terrains}

We qualitatively evaluate our dribbling controller under teleoperation on diverse terrains with different ball-terrain dynamics.
A locomotion controller must adapt when the terrain causes the feet to slip or stumble. Dribbling additionally requires that running and kicking adjust depending on how the ball interacts with the terrain, which may be very different due to the ball's lighter mass and rolling surface contact.
For example, on grass, high drag tends to slow down the rolling ball; on pavement, low drag may cause it to speed away from the robot; on gravel, the ball tends to roll, but the robot slips; on snow or sand, both ball and robot slip; on uneven and bumpy terrain, the ball changes direction unpredictably as it impacts the terrain surface.
DribbleBot is able to execute dribbling and turning motions on each terrain, tracking ball velocity commands from a human. These test terrains are illustrated in Figure \ref{fig:setup}, and the behavior is fully shown in the supplementary video.

\begin{figure*}[t]
    \vspace{0.25cm}
    \parbox{\figrasterwd}{
    \centering
    \parbox{.53\figrasterwd}{%
      \subcaptionbox{$180^{\circ}$ turn maneuver.}{\includegraphics[width=\hsize]{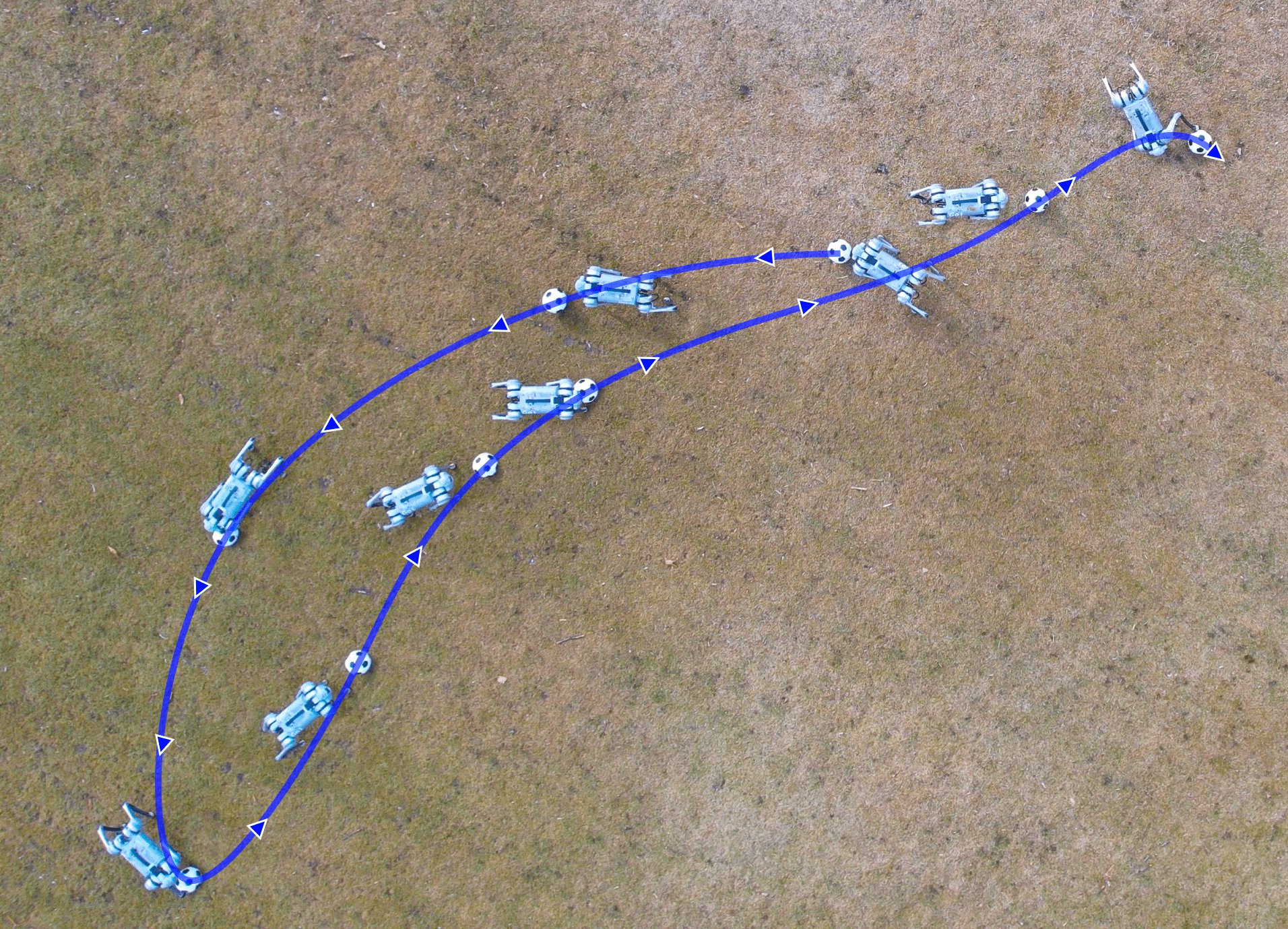}}
    }
    \hskip1em
    \parbox{.35\figrasterwd}{%
      \subcaptionbox{Ball control on tile, gravel, and bumpy moss.}{\includegraphics[clip,trim=0cm 0.8cm 0.8cm 0.8cm,width=\hsize]{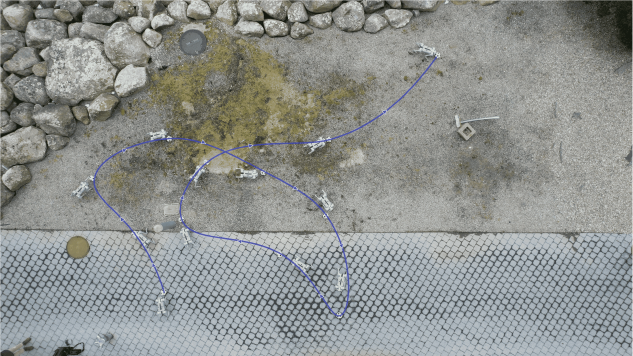}}
      \vskip0.5em
      \subcaptionbox{Dribble to goal with an evasive turn.}{\includegraphics[clip,trim=0.4cm 0.8cm 0.4cm 0.7cm,width=\hsize]{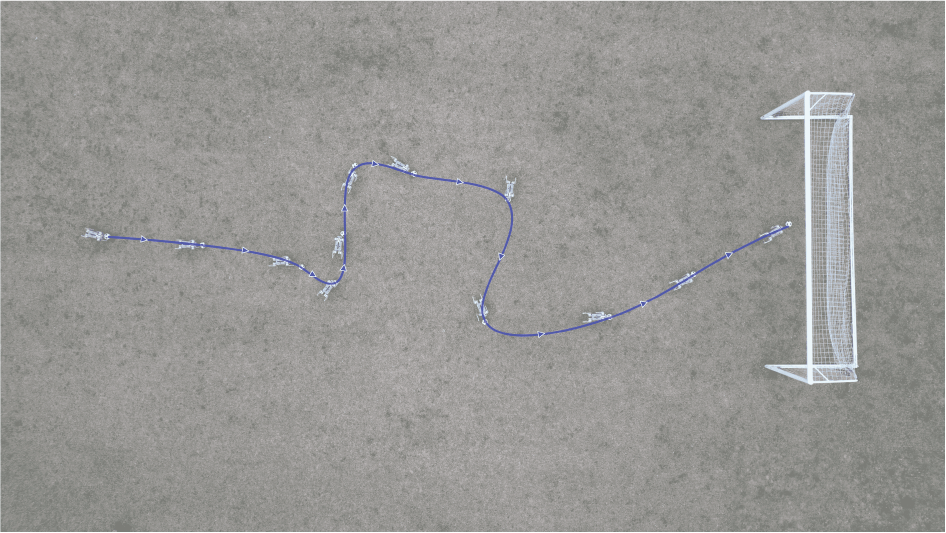}}  
    }
    }
    \caption{
    \textbf{Overhead images of DribbleBot during real-world deployment.}
    \vspace{-0.25cm}
    }\label{fig:drib_real_perf}
\end{figure*}

Because our system operates without a tether or external sensing, it is capable of manipulating the soccer ball across large outdoor spaces. To illustrate this, we collect drone footage of the robot's entire path across (a) a $180^{\circ}$ turn, (b) a series of diverse terrains including tile, gravel, and bumpy moss, and (c) a \SI{10}{\meter} run towards a soccer goal, with an evasive turning maneuver. Figure \ref{fig:drib_real_perf} shows stitched overhead photos to illustrate the real-world dribbling performance.

\subsubsection{Quantitative Results and Ablation}

We quantitatively evaluate the fully autonomous behavior of DribbleBot while executing a scripted trajectory across diverse terrains. The robot is commanded with a predetermined trajectory: dribble forward at \SI{1.5}{\meter/\second} for \SI{10}{\second}, then stop the ball for \SI{5}{\second}, then return towards the starting line at \SI{1.5}{\meter/\second} until the line is reached. We count a trial as a failure if the robot loses control of the ball, although if the loss is due to the robot falling, we allow it to recover autonomously and continue the attempt. 

As shown in Table \ref{tab:evaulation_result}, the robot executed four consecutive successful maneuvers using the full system design on tile, grass, and sand. Snow, step-down, and ramp are progressively more challenging and yield lower performance. Because the system was never exposed to steps or ramps during training, they are examples of out-of-distribution terrain for our policy. In the step-down task, the robot once fell in a pile of snow and failed to recover, and once knocked the ball far away as it fell and could not perceive it upon recovery. The ramp was traversed successfully under teleoperation,  but the robot did not make substantial forward progress in the standard experiment when the dribbling commands were pre-specified. The ball rolled back down the ramp between kicks, and the robot had to turn around to recover it. 

We also conduct an ablation study to quantify the impact of our design choices on real-world performance. We evaluate the ablated configurations under the same methodology as above: No recovery controller (\texttt{-R}); No YOLO fine-tuning (\texttt{-Y}); No custom ball drag model during training (\texttt{-D}). Ablation results show that YOLO fine-tuning (\texttt{-Y}) is critical to performance, and recovery policy (\texttt{-R}) improves one run in the challenging curb environment. The system without ball drag model (\texttt{-D}) maintains control on both grass and tile, but the video supplement shows that it dribbles substantially slower on grass despite the equal velocity command. On sand, the policy without drag model fails twice during the turning maneuver after missing a kick on the ball. This suggests that the additional robustness from the drag model may also improve response to unexpected ball motion caused by other variations like slippery terrain. 

\subsubsection{Playing with a Human and Emergent Behaviors}
A real soccer match is not played alone but with another agent who is seeking to control the ball. To understand the robot's behavior under this scenario, we explore the setting where the robot interactively plays with a human partner. 
Unlike the typical locomotion task, the task of controlling ball velocity affords the robot a high degree of freedom in its behavior, even when the user is not changing the command. Successful dribbling is not a monolithic skill: it often involves extended aperiodic movements to reach the ball, orient the body for a kick, and double back if the ball has been lost due to an unexpected perturbation, temporary visual perception failure, or control failure. We find that DribbleBot exhibits such behaviors to maintain ball control during human interaction. % \pulkit{What is the outcome of our investigation?} 

\section{RELATED WORK}

\subsection{Soccer Skills for Legged Robots}
Soccer has long been an area of interest for roboticists. The RoboCup competition, this year in its $26$th season, has attracted thousands of annual participants. RoboCup teams have implemented effective rule-based approaches to kicking, passing, and shooting in the past \cite{stone2007intelligent, veloso1998playing,friedmann2008versatile}. 

Recently, some works have applied learning to legged ball manipulation tasks in simulation \cite{peng2019mcp, muzio2022deep, xie2022learning} and in externally instrumented indoor settings \cite{shi2021circus, ji2022soccer, huang2022creating}. \cite{shi2021circus} demonstrated that a quadruped lying on its back can control and reorient a ball with its legs. A number of soccer skills, such as dribbling~\cite{peng2019mcp, peng2021amp} and juggling~\cite{xie2022learning}, have been demonstrated for physically simulated characters using reinforcement learning. \cite{bohez2022imitate} used imitation learning to perform static dribbling in the real-world indoor setting assisted by motion capture. \cite{ji2022soccer} applied a hierarchical framework to the soccer shooting task in the real world, selecting the front right foot B\'ezier curve parameters as the low-level command inputs and leveraging a real-world fine-tuning stage to improve the shooting accuracy. \cite{huang2022creating} trained a control policy for jumping to block an oncoming ball in an instrumented laboratory setting using sim-to-real reinforcement learning.  \cite{byravan2022nerf2real} learned a humanoid dribbling policy directly from onboard RGB images by rendering images from a NeRF of the target laboratory environment during simulated training.

In addition to low-level skill learning, some work has focused on learning high-level soccer play end-to-end. Notably,
\cite{liu2022motor} approaches the problems of muscle level control and long-horizon decision-making by first pretraining low-level skills using human soccer players' motion-capture video clips and then finding solutions for the multi-agent coordination goal in the low-level control space using reinforcement learning. To learn low-level skills like dribbling, \cite{liu2022motor} relies on motion capture data of human soccer players, which is not available for the quadruped form factor.

\subsection{Dynamic Object Manipulation}
Prior work has explored manipulating objects dynamically using a fixed or fully actuated base. \cite{ploeger2020high} controlled a robotic arm to blindly perform ball juggling using an open-loop policy. Another work on robotic table tennis~\cite{mulling2013learning} estimated the ball state using an extended Kalman Filter, which internally leverages a model of flight and bouncing behaviors. \cite{zeng2020tossingbot} learned a residue physics model to randomly pick up and throw a rigid object into a box. \cite{abeyruwan2022sim2real} bootstrapped a human behavior model and trained on both simulated and real data to learn a control policy for a table tennis-playing robot.

Another relevant line of work has investigated manipulating objects using a quadruped with mounted arm. \cite{mittal2022articulated} manipulated objects with a quadruped-mounted arm, coordinating the body and arm motion through a learned estimation module. \cite{ma2022combining} implemented a model-based controller to manipulate objects with a quadruped-mounted arm in a standing pose. \cite{fu2022deep} trained an end-to-end controller using reinforcement learning to perform coordinated manipulation with a quadruped-mounted arm under teleoperation and demonstrate vision-guided reaching using AprilTags. These works investigate complementary problems in the space of dynamic mobile manipulation.

\section{DISCUSSION}

DribbleBot has a number of limitations which we hope to explore and improve upon in future work. We enumerate several here, with videos of failure cases available on the project website. \textit{Slow turning response}: our system can execute sharp turns of the ball, but there is a lag between the command onset and the actual turn (Figure \ref{fig:drib_sim_perf}). \textit{Visual perception sensitivity to lighting}: We found that the visual perception module can perform poorly in bright, direct sunlight that produces glare from reflection on the ball and cameras. Fine-tuning the visual perception network with a more diverse set of outdoor images may resolve this problem. \textit{Imprecision at high speeds}: If the ball is moving too fast on low-drag terrain, or a sharp instantaneous turn is commanded at high speed, a missed attempt to stop the ball can fail, with the robot losing sight. \textit{Lack of geometry awareness}: While the robot can dribble on slippery and uneven terrains, it cannot traverse larger obstacles like steep slopes and staircases with good consistency. Moreover, it is not aware of objects in the environment, like poles and walls. Future work could incorporate more information about the environment geometry into the controller to improve ball control in cluttered and harsh settings.  

We believe there are many exciting frontiers to explore with a strong baseline for in-the-wild dribbling. Dribbling is just one component of soccer. In particular, a combination of shooting~\cite{ji2022soccer} and goalkeeping~\cite{huang2022creating} skills, as well as high-level gameplay and awareness of other agents, will be required to play a competitive game. Similarly, applying dynamic mobile manipulation for practical tasks like delivery and emergency response will require diverse skills, high-level planning, and rich world understanding. In-the-wild soccer may further be an interesting context in which to study human-robot interaction. While direct physical interaction with a legged robot is typically limited, interaction through the soccer ball as a shared medium proves rich and fun. Future work could explore how the robot is perceived by humans during play.


%%%%%%%%%%%%%%%%%%%%%%%%%%%%%%%%%%%%%%%%%%%%%%%%%%%%%%%%%%%%%%%%%%%%%%%%%%%%%%%%

%%%%%%%%%%%%%%%%%%%%%%%%%%%%%%%%%%%%%%%%%%%%%%%%%%%%%%%%%%%%%%%%%%%%%%%%%%%%%%%%

%%%%%%%%%%%%%%%%%%%%%%%%%%%%%%%%%%%%%%%%%%%%%%%%%%%%%%%%%%%%%%%%%%%%%%%%%%%%%%%%

%%%%%%%%%%%%%%%%%%%%%%%%%%%%%%%%%%%%%%%%%%%%%%%%%%%%%%%%%%%%%%%%%%%%%%%%%%%%%%%%

\section*{ACKNOWLEDGMENT}
\footnotesize{
We thank the members of the Improbable AI lab for helpful discussions and feedback. We are grateful to MIT Supercloud and the Lincoln Laboratory Supercomputing Center for providing HPC resources. This research was supported by the DARPA Machine Common Sense Program, the MIT-IBM Watson AI Lab, and the National Science Foundation under Cooperative Agreement PHY-2019786 (The NSF AI Institute for Artificial Intelligence and Fundamental Interactions, http://iaifi.org/). This research was also sponsored by the United States Air Force Research Laboratory and the United States Air Force Artificial Intelligence Accelerator and was accomplished under Cooperative Agreement Number FA8750-19-2-1000. The views and conclusions contained in this document are those of the authors and should not be interpreted as representing the official policies, either expressed or implied, of the United States Air Force or the U.S. Government. The U.S. Government is authorized to reproduce and distribute reprints for Government purposes, notwithstanding any copyright notation herein.
}

\section*{AUTHOR CONTRIBUTIONS}
\small{
\begin{itemize}%[leftmargin=*]
\item \textbf{Yandong Ji} contributed to ideation and implementation of the entire system, experimental evaluation, and writing.
\item \textbf{Gabriel B. Margolis} contributed to ideation and implementation of the entire system, experimental evaluation, and writing.
\item \textbf{Pulkit Agrawal} advised the project and contributed to its development, experimental design, and writing.
\end{itemize}
}

\bibliographystyle{IEEEtran}
\bibliography{IEEEabrv,IEEEfull}{}

\mbox{~}
\clearpage
\newpage

\section*{APPENDIX}

\section{Reward Structure}
\label{sec:reward}

\begin{table}[]
    \centering
    
    \bgroup
    \def\arraystretch{1.3}
    \caption
      {%
        Notation.
        \label{tab:definitions}%
      }
    { \scriptsize
    \begin{tabular}{rllr}
    \toprule
    % \midrule
                Parameter           &  Definition  &       Units & Dimension   \\
    \midrule
    \cline{1-4}
    \multicolumn{4}{|c|}{\textit{System Components}} \\
    \cline{1-4}
    $\pi_d$ & Dribbling Policy & - & - \\
    $\pi_r$ & Recovery Policy & - & - \\
    $\mathbf{Y}$ & YOLOv7 Network~\cite{wang2022yolov7} & - & - \\
    \cline{1-4}
    \multicolumn{4}{|c|}{\textit{Robot State}} \\
    \cline{1-4}
    $\mathbf{q}$ & Joint Angles & \SI{}{\radian} & $12$ \\
    $\dot{\mathbf{q}}$ & Joint Velocities & \SI{}{\radian/\second} & $12$ \\
    $\ddot{\mathbf{q}}$ & Joint Accelerations & \SI{}{\radian/\second} & $12$ \\
    $\pmb{\tau}$ & Joint Torques & \SI{}{\radian/\second} & $12$ \\
    $\mathbf{g}$ & Gravity Unit Vector, Body Frame & \SI{}{\meter/\second^2} & $3$ \\
    $\pmb{\psi}_t$ & Body Yaw, Global Frame & \SI{}{\radian} & $1$ \\
    $\mathbf{q}_{\text{des}}$ & Joint Position Targets & \SI{}{\radian} & $12$ \\
    $\pmb{\theta}^\mathrm{cmd}$ & Timing Reference Variables~\cite{margolis2022walktheseways} & - & $4$ \\
    $\mathbf{p}_{\mathrm{FRHip}}$ & Front Right Hip Position & \SI{}{\meter} & $3$ \\
    \cline{1-4}
    \multicolumn{4}{|c|}{\textit{Ball State}} \\
    \cline{1-4}
    $\mathbf{o}^v$ & Fisheye Camera Image & - & $400\times480$  \\
    $\mathbf{b}$ & Ball Position, Body Frame & \SI{}{\meter} & $3$ \\
    $\hat{\mathbf{b}}$ & Estimated Ball Position, Body Frame & \SI{}{\meter} & $3$ \\
    $\mathbf{v}^b$ & Ball Velocity, Global Frame & \SI{}{\meter/\second} & $2$ \\
    $\mathbf{v}^{\mathrm{cmd}}$ & Command Ball Velocity, Global Frame & \SI{}{\meter/\second} & $2$ \\
    $\psi_b$ & Direction of Ball Velocity & \SI{}{\radian} & $1$ \\
    $\psi_\mathrm{cmd}$ & Direction of Command Ball Velocity & \SI{}{\radian} & $1$ \\
    \cline{1-4}
    \multicolumn{4}{|c|}{\textit{Control Policy}} \\
    \cline{1-4}
    $\mathbf{o}$ & Policy Observation & - & $37 \times 15$ \\
    $\mathbf{a}$ & Policy Action & - & $12$ \\
    $\mathbf{c}$ & Command & - & $2$ \\
    
    \bottomrule
    \end{tabular}
    }
    \egroup
    % \vspace{-0.4cm}
    
\end{table}

\textbf{Dribbling Policy}: Table \ref{tab:reward_table} provides the reward terms for learning soccer dribbling. 
$\mathbf{v}^b$ and $\mathbf{v}^\mathrm{cmd}$ are the desired and commanded ball velocity in the global reference frame. $\psi_b$ and $\psi_\mathrm{cmd}$ are the desired and commanded direction of the ball velocity in the global reference frame, which can be directly computed from $\mathbf{v}^b$ and $\mathbf{v}^\mathrm{cmd}$. $\mathbf{b}$ is the ball position in the body frame. $\mathbf{p}_{\mathrm{FRHip}}$ is the front right hip position in the body frame. $e_\mathrm{rbcmd}$ represents the angle difference between the robot ball vector and $\psi_b$. $e_\mathrm{rbbase}$ is the angle difference between the base yaw angle and $\psi_b$. $\pmb{\kappa}$ is the target contact state $\mathbf{C}_{\mathrm{foot}}^{\mathrm{cmd}}(\pmb{\theta}^{\mathrm{cmd}},t)$, where, as in prior work~\cite{margolis2022walktheseways}, $\mathbf{f}^{\mathrm{foot}}$ is foot contact force in stance phase, 
$\mathbf{v}^{\mathrm{foot}}$ is foot velocity in swing phase, 
$\mathbf{C}_\mathrm{foot}^\mathrm{cmd}$ denotes desired foot contact sequence, and
$\pmb{\theta}^\mathrm{cmd}$ denotes timing reference variables.
$\pmb{\tau}, \mathbf{q}, \dot{\mathbf{q}}, \ddot{\mathbf{q}}$ denote joint torque, position, velocity and acceleration. $i$ is the index of joints. $\mathbf{g}^{\mathbf{xy}}_t$ denotes the gravity unit vector projected onto the robot transverse plane. $\mathbf{a}_t$ denotes the action at timestep $t$. $|\cdot|$ denotes $l^2$-norm.
The total reward at timestep $t$ is represented as $r_t=r_t^{\mathrm{pos}}\mathrm{exp}(r_t^{\mathrm{neg}})$, where $r_t^\mathrm{pos}$ and $ r_t^{\mathrm{neg}} $ represent positive task reward and negative penalizing reward respectively~\cite{ji2022concurrent}.

\textbf{Recovery Policy}: Table \ref{tab:reward_table} provides the reward terms for learning the recovery policy, inspired by~\cite{lee2019robust, smith2022legged}. Here, $\mathbf{g}_\mathrm{z}$ is the vertical component of the gravity unit vector in the body frame. When the robot is perfectly upright, $\mathbf{g}_\mathrm{z}=-1$. $\mathcal{I}$ is shorthand for an indicator variable $\mathds{1}_{\mathbf{g}_\mathrm{z}<-0.6}$ which denotes that the body height, body pose, and foot height rewards are only activated when the robot's body is nearly upright. $\mathrm{clamp}$ clamps the value of its input between $0$ and $1$.

\section{Noise Model}
\label{sec:noise_model}

\textbf{Robot Physics Randomization}: We randomize the robot's payload mass, motor strength, joint calibration, foot friction, foot restitution, and center of mass displacement. Table \ref{tab:domain_rand} provides the ranges of randomized parameters.

\textbf{Ball Physics Randomization}: We randomize the ball mass and ball-terrain drag coefficient. Table \ref{tab:domain_rand} provides the ranges of randomized parameters.

\textbf{Terrain Physics Randomization}: We did not perform substantial randomization of the terrain geometry and gravitational force while training the dribbling policy, because we observed that many such combinations make successful dribbling infeasible. However, we did train the recovery policy on randomized terrains, to enable recovery under such scenarios. To emulate rough terrain, we randomized the magnitude of perlin noise on the terrain height. To emulate sloped terrain, we applied uniform random perturbation to the gravity vector in all axes every \SI{6.0}{\second}. Table \ref{tab:domain_rand} provides the ranges of randomized parameters.

\textbf{Ball Teleportation}: We teleport the ball to a uniformly sampled random location within \SI{1.0}{\meter} at regular intervals of \SI{7.0}{\second}.

\textbf{Camera Delay}: We model the arrival time of the next observation as a Poisson distribution with mean arrival time randomized each episode between $20$\SI{}{\milli\second} and $60$\SI{}{\milli\second}.

\begin{table}
\vspace{0.3cm}
\centering
\bgroup
\def\arraystretch{1.5}
\caption{Reward terms for ball dribbling and recovery policies.}
\label{tab:reward_table}
\scriptsize
\begin{tabular}{lrr}

\cline{1-3}
\multicolumn{3}{|c|}{\textbf{Ball Dribbling Policy}} \\
\cline{1-3}
{Term}            & {Expression} & {Weight}                    \\
\hline
Projected Ball Velocity & exp\{$-\delta_{v}|\mathbf{v}^b - \mathbf{v}^\mathrm{cmd}|^2$\}    & 0.5    \\ 
Robot Ball Distance         & exp\{$-\delta_{p} |\mathbf{b}-\mathbf{p}_{\mathrm{FRHip}}|^2$\}  & 4.0  \\
Yaw Alignment   & exp\{$-\delta_{\psi}(e_{\mathrm{rbcmd}}^2+e_{\mathrm{rbbase}}^2)$\}  & 4.0
\\
Ball Velocity Norm     & exp$\{-\delta_n( |\mathbf{v}^\mathrm{cmd}|-|\mathbf{v}^b| )^2\} $        & 4.0          \\ 
Ball Velocity Angle     & 1$-\left(\psi_b-\psi_\mathrm{cmd}\right)^2/\pi^2$      & 4.0            \\  
Swing Phase Schedule   & $[1-\pmb{\kappa}]\mathrm{exp}\{-\delta_{\mathrm{cf}}|\mathbf{f}^{\mathrm{foot}}|^2\}  $     & 4.0           \\
Stance Phase Schedule    & $ \pmb{\kappa} \mathrm{exp}\{-\delta_\mathrm{cv}|\mathbf{v}^{\mathrm{foot}}_\mathrm{xy}|^2\}$     & 4.0       \\
Joint Limit Violation & $\mathds{1}_{q_i>q_{\mathrm{max} }||q_i<q_{\mathrm{min} }}$ & -10.0 \\
Joint Torque    &     $|\boldsymbol{\tau}|^2$ & -0.0001        \\
Joint Velocity & $|\dot{\mathbf{q}}|$ & -0.0001 \\
Joint Acceleration & $|\ddot{\mathbf{q}}|$ & -2.5e-7 \\
Hip/Thigh Collision & $\mathds{1}_\mathrm{collision}$ & -5.0 \\
Projected Gravity    &    $|\mathbf{g}_\mathrm{xy}|^2$ & -5.0         \\
Action Smoothing & $|\mathbf{a}_{t-1}-\mathbf{a}_t|^2$ & -0.1 \\
Action Smoothing 2 & $|\mathbf{a}_{t-2}-2\mathbf{a}_{t-1}+\mathbf{a}_{t}|^2$ & -0.1 \\

% \hline

\cline{1-3}
\multicolumn{3}{|c|}{\textbf{Recovery Policy}} \\
\cline{1-3}
{Term}            & {Expression} & {Weight}                    \\
\hline
Body Orientation    &    $(0.5 - 0.5 \mathbf{g}_\mathrm{z})^2$ &   $1.0$   \\
Body Height    &    $\mathcal{I}(1.0-\mathrm{clamp}(\frac{{h^\text{body}_{\text{target}}-h^\text{body}}}{ h^\text{body}_{\text{target}}}^2)$   &   $1.0$   \\
Body Pose &  $\mathcal{I}(1.0-\mathrm{clamp}(|q-q_\text{standing}|^2 / 20.0))$  &  $1.0$  \\
Foot Height &  $\mathcal{I}(\exp{-10 \sum_i (h^{\text{foot}}_i)^2})$  &  1.0  \\
Action & $|\mathbf{a}_t|^2$ &    $-1\mathrm{e}{-3}$   \\
Joint Torque & $|\boldsymbol{\tau}|^2$ &   $-1\mathrm{e}{-5}$   \\
\hline
\end{tabular}
\egroup
\end{table}

\begin{table}[]
    \centering
    \bgroup
    \def\arraystretch{1.3}
    \caption
      {%
        Randomization ranges for robot dynamics, ball dynamics, and commands during training.
        \label{tab:domain_rand}%
      }
    { \scriptsize
    \begin{tabular}{lrr}
    \toprule
                Dynamics Parameter           &  Range &       Units   \\
    \midrule
    \cline{1-3}
    \multicolumn{3}{|c|}{\textit{Robot Dynamics}} \\
    \cline{1-3}
    Payload Mass               & $[-1.0, 3.0]$ &  \;  \SI{}{\kilogram}  \\
    Motor Strength             &  $[90, 110]$ &  \;  \(\%\)  \\ 
    Joint Calibration             &  $[-0.02, 0.02]$ &  \;  \SI{}{\radian}  \\ 
    Robot-Terrain Friction            & $[0.40, 1.00]$ &  \;\;   --    \\
    Robot-Terrain Restitution         & $[0.00, 1.00]$ &  \;\;   --    \\
    Robot Center of Mass Displacement         & $[-0.15, 0.15]$ &  \;\;   \SI{}{\meter}    \\
    \cline{1-3}
    \multicolumn{3}{|c|}{\textit{Ball Dynamics}} \\
    \cline{1-3}
    Mass               & $[0.159, 0.254]$ &  \;  \SI{}{\kilogram}  \\
    Camera Frame Arrival Rate        & $[0.3, 0.7]$ &  \;\;   --    \\
    Teleporting Position       & $[0.0, 1.0]$ &  \;\;   \SI{}{\meter}   \\
    Perturbation Velocity       & $[0.0, 0.3]$ &  \;\;   \SI{}{\meter/\second}   \\
    Ball-Terrain Drag Coefficient       & $[0.0, 1.5]$ &  \;\;   --   \\
    \cline{1-3}
    \multicolumn{3}{|c|}{\textit{Terrain Dynamics} (Recovery Controller Only)} \\
    \cline{1-3}
    Perlin Noise Magnitude               & $[0, 10]$ &  \;  \SI{}{\centi\meter}  \\
    Gravitational Force Noise        & $[-1.0, 1.0]$ &  \;\;  \SI{}{\meter/\second^2}   \\
    \cline{1-3}
    \multicolumn{3}{|c|}{\textit{Command}} \\
    \cline{1-3}
    $\vxcmd$            & $[-1.5, 1.5]$ &  \;   \SI{}{\meter/\second}    \\
    $\vycmd$         & $[-1.5, 1.5]$ &  \;   \SI{}{\meter/\second}    \\
    \bottomrule
    \end{tabular}
    }
    \egroup
    \vspace{-0.4cm}
    
\end{table}

\section{Policy Optimization}
\label{sec:policy_details}

We used the same set of PPO hyperparameters for training the dribbling and recovery policies. Table \ref{tbl:ppo_hparams} provides these hyperparameter values. They are the same settings used in prior work for training locomotion policies on this robot~\cite{margolis2022walktheseways}.

\begin{table}
        \centering
        \raisebox{-0.0\height}{
        \small
        \begin{tabular}{lr}
        \toprule
        Hyperparameter & Value \\ [0.5ex]
        \midrule
         Discount factor & $0.99$ \\ [0.5ex]
         GAE parameter & $0.95$ \\ [0.5ex]
          Timesteps per rollout & $21$ \\ [0.5ex]
          Epochs per rollout & $5$ \\ [0.5ex]
          Minibatches per epoch & $4$ \\ [0.5ex]
          Entropy bonus ($\alpha_2$) & $0.01$ \\ [0.5ex]
          Value loss coefficient ($\alpha_1$) & $1.0$ \\ [0.5ex]
          Clip range & $0.2$ \\ [0.5ex]
          Reward normalization & yes \\ [0.5ex]
          Learning rate & $1\mathrm{e}{-3}$ \\ [0.5ex]
          \# Environments & $4096$ \\ [0.5ex]
          \# Total timesteps & $7$B \\ [0.5ex]
          Optimizer & Adam \\ [0.5ex]
         \bottomrule
        \end{tabular}
        }
        \vspace{0.45cm}
        \captionof{table}
          {%
            PPO hyperparameters. 
        \label{tbl:ppo_hparams}
          }
\end{table}

\end{document}